\documentclass[times, review, 10pt]{elsarticle}
\usepackage{geometry}
\usepackage{amssymb}
\usepackage{subfigure}
\usepackage{longtable}
\usepackage{booktabs}
\usepackage{multirow}
\usepackage{mathrsfs}
\usepackage{amsfonts}
\usepackage{amsmath}
\usepackage{array}
\usepackage{color}
\usepackage[ruled]{algorithm2e}
\usepackage{float}
\biboptions{numbers,sort&compress}
\journal{Computer}
\begin{document}
	\begin{frontmatter}
		
		\title{Three-Class Text Sentiment Analysis Based on LSTM}

		\author[firstauthor]{Yin Qixuan}
		
		\address[firstauthor]{School of Information Engineering, Zhongnan University of Economics and Law, Wuhan, 430073, China}
		
		\begin{abstract}
			
			Sentiment analysis is a critical task in natural language processing (NLP) and has widespread applications in areas such as public opinion monitoring and market research. This paper proposes a three-class sentiment analysis method based on Long Short-Term Memory (LSTM) networks, aiming to analyze the sentiment polarity (positive, neutral, negative) of Weibo comment texts. As a deep learning model, LSTM is capable of effectively handling long-distance dependencies in text, offering advantages over traditional machine learning models. By preprocessing and learning features from Weibo comment texts, the LSTM model can accurately predict the sentiment category of the comments. Experimental results show that the LSTM model performs excellently in the sentiment analysis task for Weibo comments, with an accuracy of 98.31\% and an F1 score of 98.28\%, far surpassing traditional models and other deep learning approaches. This demonstrates that LSTM can better capture the sentiment information in text, significantly improving the accuracy of sentiment classification. However, the LSTM model still has some limitations, such as high computational complexity and slower processing of long texts. Additionally, complex emotional expressions like sarcasm and humor may affect the model's performance. In the future, combining pre-trained models or improving feature engineering may further enhance the accuracy and practicality of the model. Overall, the LSTM model provides an effective solution for sentiment analysis of Weibo comments.\cite{Zhang2010}
			
		\end{abstract}
		
		
	\end{frontmatter}
	
	\section{Introduction}
	With the rapid development of computer technology and the internet, information dissemination has increasingly become dominated by the internet, with social networks serving as key platforms for users to express opinions and emotions. On social media and short video platforms, user engagement has been continuously increasing. By the end of the first quarter of 2023, the number of monthly active users on Weibo reached 593 million, with 255 million daily active users. Especially during sudden public events, user interaction frequency on Weibo has significantly increased, forming a public opinion sphere in a short time. Moreover, Weibo, as an amplifier for information dissemination, has also enhanced the influence of online public opinion.
	
	However, with the vast amount of comment data available, it is becoming increasingly difficult to manually analyze the potential direction and impact of public opinion on hot topics. Therefore, extracting key information from billions of comment texts and accurately analyzing users' emotional trends to provide refined public opinion guidance has become a significant challenge in the current research field.
	
	Text sentiment analysis technology can extract valuable information from massive social media data, greatly improving the efficiency of user sentiment analysis. Although there have been significant advances in sentiment analysis research that combines deep learning in recent years, sentiment analysis of social media comment data still faces three major challenges:
	1. Diversity of comment forms: Comments on different social platforms often present in multimodal forms, requiring consideration of the fusion of different modal features.
	2. Informality of sentences and fast updates of new words: User comments are often informal, with low grammatical standards, and frequently use emerging popular terms, making sentiment analysis methods based on dictionaries or rules difficult to handle.
	3. Complexity of sentiment word parts of speech: In Chinese corpora, a significant proportion of sentiment words are contained in verbs and nouns, but low-frequency key noun sentiment words are often overlooked, making it difficult for sentiment word extraction or feature extraction models to capture the core emotional information in sentences.
	
	This paper conducts experiments on Weibo comment texts, analyzing users' attention to hot public opinion events and their emotional attitudes. To effectively address the above challenges, this paper proposes a three-class sentiment analysis method based on Long Short-Term Memory (LSTM). LSTM captures the contextual semantics of texts and performs sentiment classification tasks. At the same time, the LSTM model demonstrates high accuracy and robustness in the comment sentiment classification task. This research not only validates the effectiveness of combining multimodal features but also provides a reference for further exploration in the field of sentiment analysis.
	
	\section{Related Work}
	
	In recent years, with the rapid development of the internet, the frequency of emotionally inclined texts on the web has gradually increased, making text sentiment analysis a hot research direction in the field of Natural Language Processing (NLP). For text sentiment analysis, research methods are mainly divided into three categories: methods based on sentiment dictionaries/rules, methods based on traditional machine learning, and methods based on deep learning.
	
	\subsection{Sentiment Analysis Methods Based on Sentiment Dictionaries/Rules }
	Sentiment analysis methods based on sentiment dictionaries are an early technology in the field of sentiment analysis. The core idea is to construct or use predefined sentiment dictionaries to analyze and compare the sentiment words in the input text, thereby predicting the sentiment polarity of the sentence. For example, the Chinese dictionary HowNet and the English dictionary WordNet are often used for the extension and optimization of sentiment dictionaries. However, the quality of sentiment dictionaries directly impacts the accuracy and stability of the model. Researchers have sought to improve classification performance by designing sentiment dictionaries from large-scale text data. For instance, some studies have constructed large-scale Chinese sentiment dictionaries using simple statistical methods, while others have optimized dictionaries by extending them with semantic relationships or domain-specific sentiment words based on existing dictionaries.  
	Despite this, the applicability of sentiment dictionaries has significant limitations. For example, the meaning of sentiment words may vary in different contexts, leading to a decrease in classification accuracy. To address this issue, researchers have tried to combine sentiment dictionary methods with supervised learning. For instance, some studies have expanded sentiment dictionaries and combined them with self-supervised learning to train sentiment classifiers, significantly improving the accuracy of sentiment classification for ambiguous text. However, due to the difficulty of sentiment dictionaries in adapting to the rapid evolution of new vocabulary, their standalone application fails to meet the demands of the big data era, and research has gradually shifted towards traditional machine learning methods.\cite{Liang2014,Wang2007}
	
	\subsection{Sentiment Analysis Methods Based on Traditional Machine Learning}
	
	Traditional machine learning methods learn and extract text features through algorithms to determine the sentiment polarity of the text. These methods effectively avoid some of the drawbacks of sentiment dictionary-based approaches and have strong adaptability. Commonly used machine learning algorithms include Support Vector Machine (SVM), Naive Bayes, Decision Trees, K-Nearest Neighbors (KNN), and Random Forest, among others.  
	In research, a multimodal joint sentiment topic model based on Weibo data (MJST) introduces emojis and user personality features. It analyzes users' hidden emotional characteristics and sentiment topic features using the Latent Dirichlet Allocation (LDA) model, which significantly improves the performance of unsupervised learning compared to traditional methods. Additionally, some studies have applied TF-IDF feature selection combined with the SVM model for sentiment classification of Tibetan corpus, enriching the research on Tibetan sentiment analysis.  
	Although traditional machine learning methods outperform sentiment dictionary approaches in terms of performance, they suffer from poor transferability and high dependence on domain knowledge and manual labeling. With the development of deep learning technologies, sentiment analysis has gradually shifted towards deep learning methods.\cite{Liu2017,SocialMedia2013}
	
	\subsection{ Sentiment Analysis Methods Based on Deep Learning}
	Deep learning achieves feature extraction and learning from data through multi-layer neural networks, and in recent years, it has shown significant advantages in text sentiment analysis. Representative models include Convolutional Neural Networks (CNN) and Recurrent Neural Networks (RNN), among others.
	
	Research has found that introducing attention mechanisms into deep learning models can significantly enhance the model's ability to identify key emotional information in text. For example, one study combined attention mechanisms with dynamic CNN and Bidirectional Gated Recurrent Units (BiGRU) to strengthen the model's ability to extract emotional features from text. Another study incorporated attention mechanisms into RNNs to further improve sentiment classification accuracy.
	
	In recent years, pre-trained models such as BERT have driven the development of text sentiment analysis. Pre-trained models improve sentiment analysis performance by pre-training general models on large-scale unlabeled corpora and then fine-tuning them for specific tasks. For example, combining BERT with LSTM allows for more fine-grained sentiment classification. Additionally, some studies have used BERT?s dynamic representations of Weibo text, extracting contextual features and processing them with Bidirectional LSTM and attention mechanisms, while CNN captures local sentiment features to effectively perform sentiment polarity classification. To further enhance model performance, other studies have integrated semantic information from different hidden layers of BERT to generate richer feature vectors, thereby improving the model's ability to recognize sentiment tendencies.
	
	In summary, deep learning-based sentiment analysis methods have progressively become mainstream, bringing new breakthroughs to sentiment analysis research in the realm of social networks. The introduction of pre-trained models has further expanded the boundaries of sentiment analysis technology.\cite{Wang2014,Xie2012}
	
	\section{Model Design and Construction}
	
	\subsection{LSTM Model Architecture}
	Long Short-Term Memory (LSTM) networks are a type of Recurrent Neural Network (RNN) well-suited for processing sequential data, capable of capturing long-term dependencies and sequence information within text data. Widely applied in natural language processing tasks, this paper opts to use an LSTM model independently for feature extraction and sentiment analysis on textual data, effectively modeling the sequential information of texts. The LSTM model primarily consists of an input layer, LSTM layer, and output layer, as illustrated in Figure 1.
	
	The processed text sequences—having undergone tokenization and vectorization—are fed into the LSTM network. In the input layer, text data is transformed into fixed-length word vector sequences, serving as inputs to the LSTM. Through its unique gating mechanisms—the input gate, forget gate, and output gate—the LSTM network processes sequential data, effectively capturing significant contextual and temporal information within texts while avoiding the vanishing or exploding gradient problems common in traditional RNNs.
	
	The output from the LSTM layer contains high-dimensional sequence features, which are then passed through a fully connected layer to further extract deep semantic information from the text. Finally, a Softmax classifier maps these features to a probability distribution across categories, completing the classification prediction for the sentiment analysis task. This overall model architecture is straightforward yet powerful, leveraging the advantages of LSTMs in sequence modeling to effectively mine emotional features from text data.
	
	\subsection{Input Layer Structure}
	The input layer of the LSTM model is responsible for converting raw text data into numerical vector representations that can be processed by the subsequent network layers. The design of the input layer involves three key steps: text preprocessing, word embedding representation, and input sequence construction.
	
	Text Preprocessing: Raw text data undergoes tokenization, dividing it into individual words or subword units. Stopwords, special symbols, and redundant characters are removed to ensure the purity of the text data. Tokenized texts are then indexed, converting them into integer indices according to a vocabulary list.
	
	Word Embedding Representation: Word Embedding technology, such as Word2Vec, GloVe, or randomly initialized embedding layers, maps the text data into continuous vector representations of fixed dimensions. Each input word is mapped into a low-dimensional word vector space where word vectors can capture semantic relationships and similarities between words. An embedding layer converts the indexed words into a word vector matrix with the shape (batch\_size, sequence\_length, embedding\_dim).
	
	Input Sequence Construction: Text data is standardized to a uniform length by setting sequence\_length, padding shorter texts, and truncating longer ones. After passing through the embedding layer, the input text sequences form tensors of shape batch\_size,sequence\_length,embedding\_dim, which serve as inputs to the LSTM layer. Through the input layer's processing, raw text data is efficiently converted into numerical sequence representations suitable for LSTM network modeling, providing a foundation for capturing sequence features.
	
	\subsection{LSTM Layer}
	
	RNNs are particularly suited for handling linear sequence data, incorporating a recurrent structure within neural networks so that the current hidden layer is influenced not only by the input layer but also by the previous time step. To address the issues of vanishing and exploding gradients, LSTMs introduce storage and memory functions, controlling information flow through gating mechanisms, selectively forgetting some information. An LSTM unit includes an input gate, forget gate, output gate, and memory cell, as depicted in Figure 1.
	
	\begin{figure}[H]
		\begin{center}
			\includegraphics[width=0.8\linewidth]{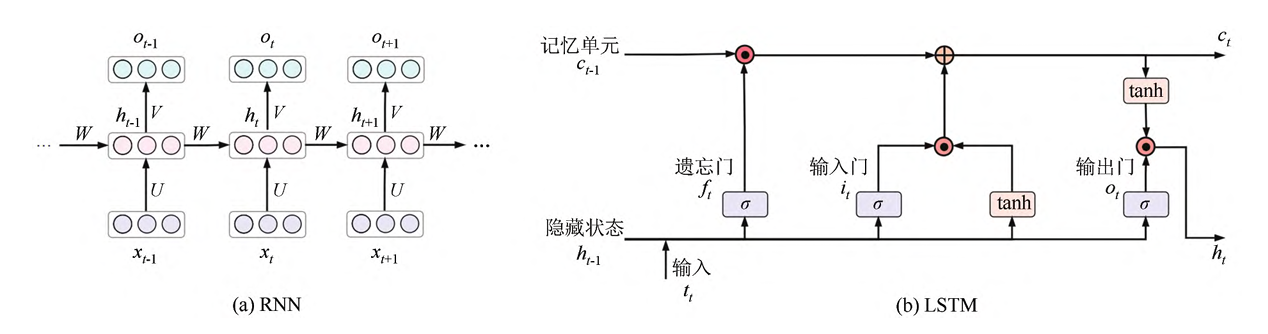}
		\end{center}
		\caption{Diagram of RNN and LSTM cell structures}
	\end{figure}
	
	Forget Gate: Determines what information should be discarded from the current memory cell. The formula, translated into R-like syntax, is:
	
	\begin{equation}
		f_t = \sigma(W_f \cdot [h_{t-1}, x_t] + b_f)
	\end{equation}
	
	\begin{figure}[H]
		\begin{center}
			\includegraphics[width=0.8\linewidth]{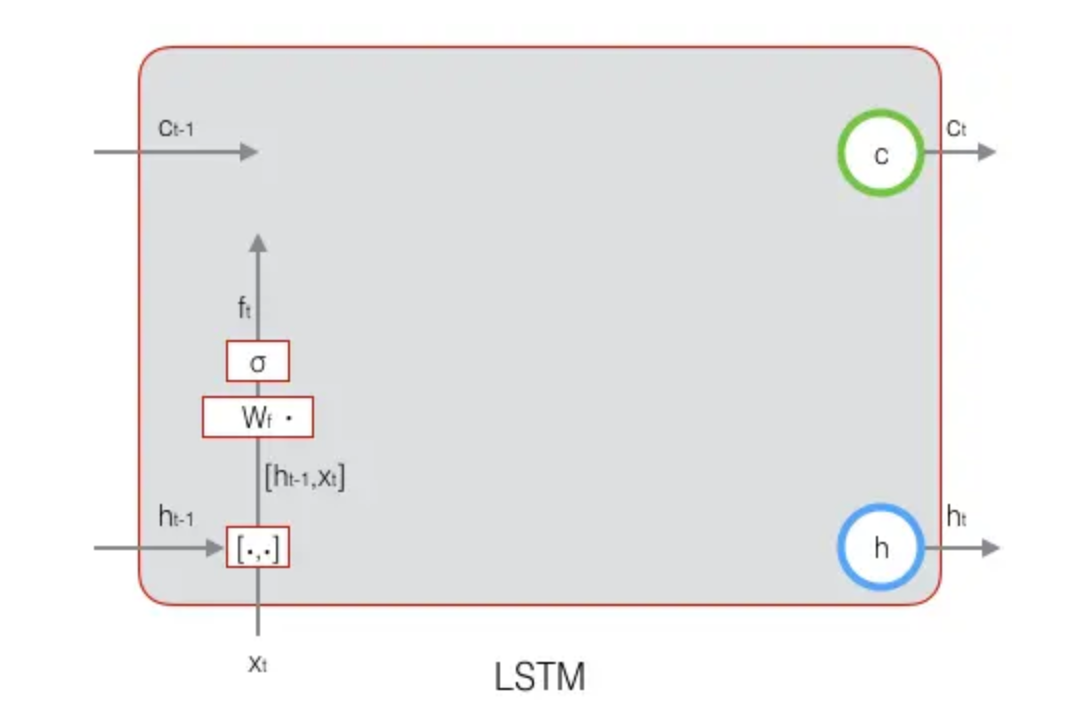}
		\end{center}
		\caption{Forget Gate}
	\end{figure}
	
	Input Gate: Decides what new information should be stored in the current memory cell. The formulas, in R-like syntax, are:
	
	\begin{equation}
		i_t = \sigma(W_i \cdot [h_{t-1}, x_t] + b_i) 
	\end{equation}
	\begin{equation}
		\tilde{c}_t = \tanh(W_c \cdot [h_{t-1}, x_t] + b_c)
	\end{equation}
	
	Update Memory Cell: Updates the state of the current memory cell. The formula, in R-like syntax, is:
	
	\begin{equation}
		c_t = f_t \times c_{t-1} + i_t \times \tilde{c}_t
	\end{equation}
	
	\begin{figure}[H]
		\begin{center}
			\includegraphics[width=0.8\linewidth]{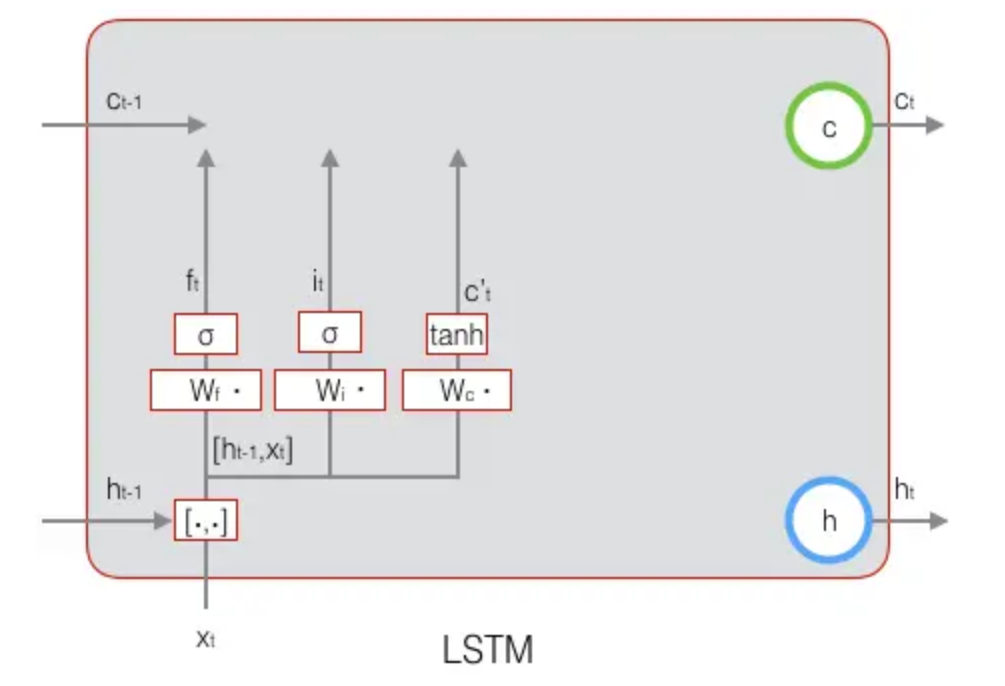}
		\end{center}
		\caption{~ct}
	\end{figure}
	
	Output Gate: Determines what part of the current memory cell should be output. The formulas, in R-like syntax, are:
	
	\begin{equation}
		O_t = \sigma(W_o \cdot [h_{t-1}, x_t] + b_o) 
	\end{equation}
	\begin{equation}
		h_t = O_t \times \tanh(c_t)
	\end{equation}
	
	\begin{figure}[H]
		\begin{center}
			\includegraphics[width=0.8\linewidth]{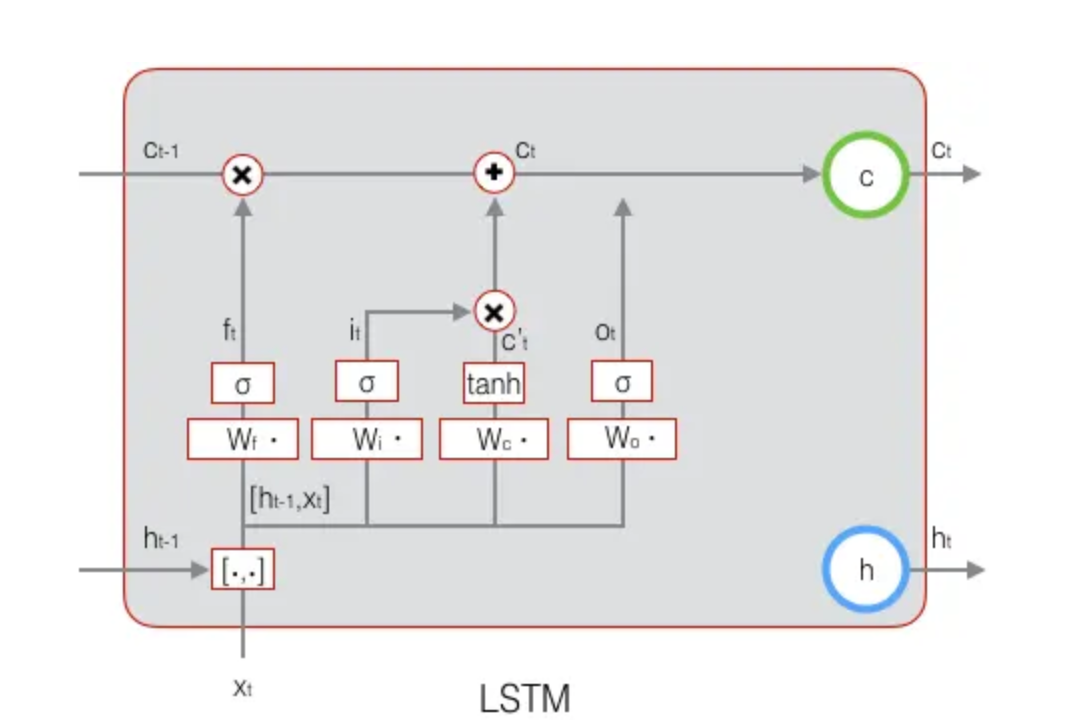}
		\end{center}
		\caption{Output Gate}
		
	\end{figure}
	\begin{figure}[H]
		\begin{center}
			\includegraphics[width=0.8\linewidth]{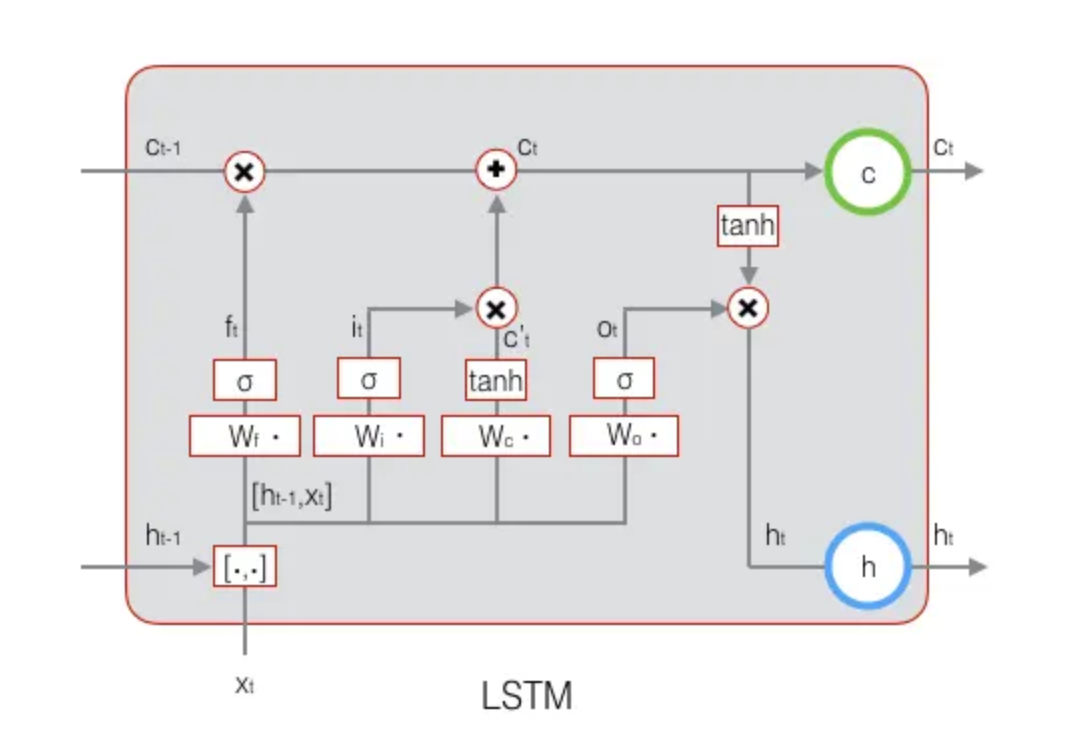}
		\end{center}
		\caption{LSTM}
	\end{figure}
	
	\subsection{Output Layer}
	The output layer feeds the feature vectors produced by the LSTM model into a fully connected layer for dimensionality reduction. The fully connected layer then outputs the sentiment label categories, thereby completing the sentiment classification task. Finally, the Softmax function is utilized to normalize the outputs, resulting in the predicted sentiment polarity for the review texts.
	
	\section{Experiments and Results Analysis}
	In this section, we first introduce our experiment settings including datasets, evaluation metrics and implementation details. Then we conduct a comparative analysis against state-of-the-art methods on various benchmark datasets. Finally, we perform the self-evaluations and ablation studies for our proposed model.
	
	\subsection{Data Preparation}
	The dataset used in this study consists of Weibo comment texts that were collected through web scraping, totaling 21,091 samples: 8,703 negative, 4,355 neutral, and 8,033 positive comments. To ensure the validity of the data, preprocessing steps were applied to the text data. This involved using regular expressions to remove punctuation, spaces, URLs, and Weibo-specific elements such as topics marked with ''\#\#'' and usernames prefixed with ''@''—all of which are irrelevant to sentiment analysis.

	\subsection{Experimental Environment Setup}
	The experimets were conducted using the PyTorch open-source deep learning framework on the AutoDL cloud environment(https://www.autodl.com/home).The specific experimental environment settings are shown in Table 1.
	
	\begin{table}[H]
		\centering
		\caption{Experimental environment}
		\begin{tabular}{ll}
			\toprule
			\multicolumn{2}{c}{\textbf{Hardware environment}} \\
			\midrule
			Operating system & Windows 10 64bit \\
			CPU     & 14 vCPU Intel(R) Xeon(R) Platinum 8362 CPU @ 2.80GHz \\
			GPU     & RTX 3090 (24GB) $\times$ 1 \\
			\midrule
			\multicolumn{2}{c}{\textbf{Software environment}} \\
			\midrule
			PyTorch & 1.11.0 (CUDA 11.3) \\
			Python  & 3.8 (Ubuntu 20.04) \\
			Pandas  & 2.0.3 \\
			NumPy   & 1.23.1 \\
			scikit-learn & 1.3.2 \\
			Gensim  & 4.3.3 \\
			TensorFlow & 2.9.0 \\
			Jieba   & 0.42.1 \\
			\bottomrule
		\end{tabular}
	\end{table}
	
	\subsection{Parameter Settings}
	
	\begin{itemize}
		\item \textbf{\texttt{vocab\_dim = 100}}: This parameter sets the dimensionality of the word vectors, mapping each word to a fixed-length vector. Smaller dimensions (e.g., 50--100) result in faster training and lower memory usage but capture limited semantic information. Larger dimensions (e.g., 300) can capture richer semantic information but require longer training times. This is commonly used in word embedding models like Word2Vec and GloVe.
		
		\item \textbf{\texttt{n\_iterations = 10}}: This specifies the number of iterations for model training. More iterations can improve the quality of word vectors but also increase computational time.
		
		\item \textbf{\texttt{n\_exposures = 10}}: This sets the word frequency threshold, retaining only words that appear more than 10 times. Filtering low-frequency words reduces noise. This parameter typically corresponds to the \texttt{min\_count} parameter in Word2Vec.
		
		\item \textbf{\texttt{window\_size = 7}}: This controls the size of the context window, considering up to 7 words to the left and right of a target word as context. A smaller window captures local relationships, while a larger window captures broader contexts.
		
		\item \textbf{\texttt{n\_epoch = 4}}: This sets the number of epochs for model training. Each epoch represents one complete pass through the entire dataset. Increasing the number of epochs can enhance model performance but may introduce overfitting risks.
		
		\item \textbf{\texttt{input\_length = 100} and \texttt{maxlen = 100}}: These parameters ensure that input sequences are fixed at length 100. Shorter texts are padded with zeros to reach the desired length, while longer texts are truncated to fit. This maintains consistency in input data, facilitating efficient computation for models like LSTM and CNN. These parameters are commonly used in Keras's Embedding layer and the \texttt{pad\_sequences} function.
	\end{itemize}
	
	\subsection{Evaluation Metrics}
	
	In this experiment, accuracy, precision, recall, and the F1 score are used as evaluation metrics. The confusion matrix for the prediction results is shown in Table~\ref{tab:confusion_matrix}.
	
	\begin{itemize}
		\item \textbf{TP (True Positive)}: Indicates cases where both the predicted sentiment and the actual sentiment are positive.
		\item \textbf{TN (True Negative)}: Indicates cases where both the predicted sentiment and the actual sentiment are negative.
		\item \textbf{FP (False Positive)}: Indicates cases where the actual sentiment is negative but the model predicts it as positive.
		\item \textbf{FN (False Negative)}: Indicates cases where the actual sentiment is positive but the model predicts it as negative.
	\end{itemize}
	
	\begin{table}[H]
		\centering
		\caption{Confusion Matrix}
		\label{tab:confusion_matrix}
		\begin{tabular}{l|cc}
			\toprule
			& \multicolumn{2}{c}{Predicted Result} \\
			Actual Result & Positive & Negative \\
			\midrule
			Positive & TP & FN \\
			Negative & FP & TN \\
			\bottomrule
		\end{tabular}
	\end{table}
	
	Accuracy represents the proportion of correctly predicted data out of all predictions. The formula for accuracy is:
	
	\begin{equation}
		A = \frac{TP + TN}{TP + TN + FP + FN}
	\end{equation}

	Precision represents the proportion of correctly predicted positive instances out of all instances predicted as positive. The formula for precision is:
	
	\begin{equation}
		P = \frac{TP}{TP + FP}
	\end{equation}
	
	Recall represents the proportion of correctly predicted positive instances out of all actual positive instances. The formula for recall is:
	
	\begin{equation}
		R = \frac{TP}{TP + FN}
	\end{equation}
	
	The F1 score is the harmonic mean of precision and recall, giving equal weight to both. The formula for the F1 score is:
	
	\begin{equation}
		F1 = \frac{2 \cdot P \cdot R}{P + R}
	\end{equation}
	
	\subsection{Results Analysis}

	In the comparative experiments, this study selected four traditional machine learning baseline models (KNN, Random Forest, Logistic Regression, Naive Bayes), two deep learning baseline models (CNN, RNN), and two deep learning ensemble models that used the same dataset as this study (ERNIE-DAM, FastText-BGRU). These models were compared with the LSTM model used in this study. The specific results are described as follows:
	
	\begin{table}[htbp]
		\centering
		\caption{Model Comparison Experimental Results}
		\label{tab:model_comparison}
		\begin{tabular}{lcccc}
			\toprule
			Model & Accuracy (\%) & Precision (\%) & Recall (\%) & F1 Score (\%) \\
			\midrule
			KNN & 55.07 & 55.65 & 69.46 & 45.75 \\
			Random Forest & 74.44 & 86.96 & 57.51 & 69.23 \\
			Logistic Regression & 83.23 & 83.12 & 83.38 & 83.26 \\
			Naive Bayes & 83.13 & 83.10 & 83.24 & 83.10 \\
			CNN & 84.80 & — & 84.90 & 84.90 \\
			RNN & 86.56 & 86.53 & 86.63 & 86.58 \\
			LSTM & 98.31 & 99.98 & 96.63 & 98.28 \\
			\bottomrule
		\end{tabular}
	\end{table}
	
	From the experimental results, we can conclude the following:
	
	\begin{enumerate}
		\item \textbf{Superiority of Deep Learning Models}:
		- Deep learning models can autonomously learn feature representations without requiring manual feature design and selection. They can automatically learn high-level, abstract features from raw data, better capturing semantic and contextual information. Traditional machine learning models typically require manual extraction and selection of appropriate features to represent text, which is a complex and labor-intensive process. Therefore, under general conditions, classification performance using deep learning algorithms outperforms traditional machine learning algorithms. In this experiment, the LSTM model showed improvements of 13\% to 16\% across all metrics compared to the best-performing traditional machine learning model.
		
		\item \textbf{Progressive Improvement in Deep Learning Baseline Models}:
		- Among the deep learning baseline models, performance improved progressively from CNN to RNN and finally to LSTM.
		- CNN primarily focuses on local features and lacks comprehensive consideration of context.
		- RNN has memory capabilities, allowing it to make predictions based on previous information. For tasks like sentiment classification, long-term context information is often more important. Thus, RNN's sentiment classification performance was 2\% higher than CNN in the comparative experiments.
		- LSTM is an optimized variant of RNN with memory cells, input gates, forget gates, and other gating mechanisms, providing stronger expressive power and better maintenance of long-term dependencies. Therefore, in the comparative experiments, LSTM's sentiment classification results were approximately 6\% higher than RNN.
	\end{enumerate}
	
	\section{Conclusion}
	This paper thoroughly examines the application of LSTM (Long Short-Term Memory) models in Chinese sentiment analysis. Experimental results show that by capturing the contextual information of samples, the LSTM model significantly enhances the performance of sentiment classification. Comparative experiments demonstrate that our proposed model outperforms various classical traditional machine learning models, thereby validating its effectiveness and feasibility to some extent.
	
	Future research can further address ambiguous comments. Specifically, for comments that contain multiple emotions, a multi-label classification approach can be employed to describe the complex emotional composition within these comments. By integrating contextual information, this method can better understand the different emotional contexts, thereby improving model performance and enhancing its generalization capability.

	\quad
	
	\bibliography{NLP}
	
\end{document}